\documentclass[lettersize,journal]{IEEEtran}
\usepackage{amsmath,amsfonts}
\usepackage{array}
\usepackage[caption=false,font=normalsize,labelfont=sf,textfont=sf]{subfig}
\usepackage{textcomp}
\usepackage{stfloats}
\usepackage{url}
\usepackage{verbatim}
\usepackage{graphicx}
\usepackage{algorithmicx}
\usepackage{algpseudocode}
\usepackage{amssymb}
\usepackage{algorithm}
\usepackage{makecell}
\usepackage{float}
\usepackage{multirow}
\usepackage{array}
\usepackage{pifont}
\usepackage{booktabs}
\usepackage[capitalize]{cleveref}

\usepackage{amsmath}
\usepackage{tikz}
\newcommand*{\circled}[1]{\lower.7ex\hbox{\tikz\draw (0pt, 0pt)%
    circle (.5em) node {\makebox[1em][c]{\small #1}};}}
\def\BibTeX{{\rm B\kern-.05em{\sc i\kern-.025em b}\kern-.08em
    T\kern-.1667em\lower.7ex\hbox{E}\kern-.125emX}}
\usepackage{balance}
\begin{document}
\title{Reliable Federated Disentangling Network for Non-IID Domain Feature}
\author{Meng Wang, Kai Yu, Chun-Mei Feng, Yiming Qian, Ke Zou, Lianyu Wang, Rick Siow Mong Goh, \\Yong Liu, and Huazhu Fu, \IEEEmembership{Senior member, IEEE}
\thanks{This study was supported in part by Huazhu Fu’s A*STAR Central Research Fund, and Career Development Fund (C222812010), A*STAR Advanced Manufacturing and Engineering~(AME) Programmatic Fund~(A20H4b0141), and the Project of Natural Science Foundation of Zhejiang Province~(LQ22F010003). Meng Wang and Kai Yu contributed equally to this work. Corresponding authors: Huazhu Fu.}
\thanks{Meng~Wang, Kai~Yu, Chun-Mei~Feng, Yiming~Qian, Rick~Siow~Mong~Goh, and Yong~Liu are with the Institute of High Performance Computing (IHPC), Agency for Science, Technology and Research (A*STAR), 1 Fusionopolis Way, \#16-16 Connexis, Singapore 138632, Republic of Singapore}
\thanks{Ke~Zou is with the National Key Laboratory of Fundamental Science on Synthetic Vision and the College of Computer Science, Sichuan University, Sichuan 610065, China.}
\thanks{Lianyu~Wang is with the College of Computer Science and Technology, Nanjing University of Aeronautics and Astronautics, Jiangsu 211100, China.}
\thanks{Huazhu~Fu is with the Institute of High Performance Computing (IHPC), Agency for Science, Technology and Research (A*STAR), 1 Fusionopolis Way, \#16-16 Connexis, Singapore 138632, Republic of Singapore~(e-mail: liuyong@ihpc.a-star.edu.sg, hzfu@ieee.org).}
}


\maketitle

\begin{abstract}
Federated learning (FL), as an effective decentralized distributed learning approach, enables multiple institutions to jointly train a model without sharing their local data. However, the domain feature shift caused by different acquisition devices/clients substantially degrades the performance of the FL model. Furthermore, most existing FL approaches aim to improve accuracy without considering reliability (e.g., confidence or uncertainty). The predictions are thus unreliable when deployed in safety-critical applications. 
Therefore, aiming at improving the performance of FL in non-Domain feature issues while enabling the model more reliable.
In this paper, we propose a novel reliable federated disentangling network, termed RFedDis, which utilizes feature disentangling to enable the ability to capture the global domain-invariant cross-client representation and preserve local client-specific feature learning.
Meanwhile, to effectively integrate the decoupled features, an uncertainty-aware decision fusion is also introduced to guide the network for dynamically integrating the decoupled features at the evidence level, while producing a reliable prediction with an estimated uncertainty. 
To the best of our knowledge, our proposed RFedDis is the first work to develop an FL approach based on evidential uncertainty combined with feature disentangling, which enhances the performance and reliability of FL in non-IID domain features. 
Extensive experimental results show that our proposed RFedDis provides outstanding performance with a high degree of reliability as compared to other state-of-the-art FL approaches. 
\end{abstract}


\section{Introduction}
\IEEEPARstart{F}{ederated} learning (FL) enables a collaborative framework for training a model by multiple clients without sharing data~\cite{FL}. FedAvg is a widely-used FL framework~\cite{FedAvg} that uses a model averaging scheme where each client trains the model independently and sends the trained model to a central server. 
The central server averages the local models from clients to aggregate into a global model. FedAvg has been a popular framework famous for its effectiveness and simplicity, inspiring a number of follow-up works~\cite{FedProx,Moon, FedRep}. However, the performance of FedAvg degrades when applied to the non-IID data, where individual clients have different label distribution but with the same data feature distribution. 
Some existing studies~\cite{FedProx,Moon,FedRep,FedDC,SCAFFOLD,FedMRI} have attempted to improve FedAvg on non-IID settings. However, these methods achieve sub-optimal performance on more challenging non-IID feature tasks.
For example, individual clients have different data feature distributions, caused by different collection devices, environments, and quality, which results in large inter-domain differences across clients. These are more challenging scenarios in the real-world deployment of FL.
Therefore, it is crucial to address the non-IID domain feature, which is one of the main focuses of this paper. Currently, the common FL algorithms for non-IID can be roughly summarized into three strategies~\cite{Zhao2018,Li2020}:
1) mitigating the drift of the global and local models by improving the objective function;
2) aggregation scheme in server with introducing measurements and  constraints;
and 3) personalized FL method sharing part of the client's network with the server for aggregation and keeping the rest of the network in the local client. 
Indeed, these approaches do achieve encouraging performance on their respective tasks.
However, these methods attempt to improve performance at the model level without exploring non-IID features in the feature space. Therefore, these methods cannot make good use of the correlation of global domain-invariant and local domain-specific features, which also leads to the limited performance of these methods on FL tasks with large differences in feature distribution between clients.

On the other hand, most existing FL algorithms focus on improving the performance~\cite{F_NonIID,F_NonIID_kairouz}, while ignoring the trustworthiness and reliability of the prediction, which may cause these models are thus unreliable and unstable for the safety-critical situations. Providing high levels of uncertainty scores for incorrect predictions is critical to the final decision~\cite{MC_Drop,Abdar2021}, which allows the system to make more reliable decisions and potentially avoid disasters from out-of-distribution (OOD) samples, such as anomaly decisions, autopilot obstacle avoidance errors, and medical imaging screening accidents. 
Some studies have been proposed to explore the field of model uncertainty, e.g., the deterministic method~\cite{Deterministic}, Bayesian method~\cite{MC_Drop,FirstUncertainty}, and ensemble-based method~\cite{EnsembleUncertainty}. These uncertainty methods only work well on centralized machine learning (ML), however, are no longer effective for FL settings, mainly due to the inherent distinction in the way how FL and ML learn from data.

To improve the non-IID domain feature performance while estimating the uncertainty for FL, we propose a novel reliable federated disentangling network, named RFedDis.
Specifically, a feature disentangling-based FL framework is proposed to guide the model to learn generic domain-invariant cross-client representations and preserve the local domain-specific features learning in the feature space. 
However, how to perform an effective fusion of the decoupled features is also a challenge, which is directly related to the final performance of FL in non-IID domain feature tasks.
Therefore, to this end, we also develop an uncertainty-aware decision fusion based on the subjective logical (SL) evidential theory to conduct dynamic decision fusion of the decoupled features while producing an uncertainty evaluation for the final decision, which improves the performance of FL in non-IID domain feature and enhances the reliability of the deployed FL model.
\textbf{Our main contributions are highlighted as follows:}
\begin{itemize}
	\setlength{\itemsep}{0pt}
	\setlength{\parsep}{-2pt}
	\setlength{\parskip}{-0pt}
	\setlength{\leftmargin}{-15pt}
\item  A reliable federated disentangling network (RFedDis) is proposed for improving the performance of FL on non-IID domain issues while enhancing the reliability of the final decision. To the best of our knowledge, our RFedDis is the first work to introduce evidential uncertainty combined with disentangled feature learning into the FL framework. 
\item A federated disentangling is designed to effectively guide the model to capture the global domain-invariant cross-client information and enhance client-specific feature learning.
Our federated disentangling approach can guide the model to learn a generalized representation globally while preserving the local client-specific feature information.
\item  An Uncertainty-aware decision fusion is introduced, which can effectively guide the network to dynamically integrate the decoupled features at the evidence level. Meanwhile, an uncertainty score is also generated to evaluate the confidence for the prediction without performance loss. 
\item  The extensive experiments on three non-IID datasets are conducted, \textit{i.e.}, OfficeCaltech10 (four domain sources), Digits (five domain sources), and DomainNet (six domain sources). Our RFedDis outperforms the state-of-the-art FL approaches. In addition, with the estimated uncertainty, our method could produce a reliable prediction and potentially avoid disasters from OOD samples. The code will be released on: \url{https://github.com/LooKing9218/RFedDis}
\end{itemize}

\section{Related Work}
\label{sec:rela}
\subsection{Federated learning on non-IID data}
Recently, FL as a distributed computing architecture without the needs of the local side to share data, has become a hot research topic~\cite{FederatedCOVID,FederatedProblems,Li2020FL}. 
One of the most critical problems holding back the deployment of FL in real-world scenarios is the feature degradation from non-IID between clients. Previous works~\cite{FedProx,Moon,FedDyn,FedRep,FedDC} attempting to solve the non-IID problem have largely focused on label distribution skewing, where the FL dataset is assumed to be an unbalanced partitioning of a homogeneous dataset.
Although these methods have achieved promising performance on non-IID label settings, their performance drops in the non-IID domain feature tasks. Recently, some studies have also been conducted on FL for non-IID domain feature settings. For example, FedBN~\cite{FedBN} mitigates the non-IID domain feature problem by keeping the local batch normalization parameters out of the global model aggregation. 
\cite{FedMRI} proposed a speciﬁcity-preserving FL algorithm for MR image reconstruction termed FedMRI, which learns a generic globally shared encoder while providing a customer-specific decoder for local reconstruction. 
There are also some works to incorporate the problem of domain generalization into FL, such as FedADG~\cite{FedADG} introduces a joint adversarial learning approach to address the problem that models trained on multiple source domains may have poor generalization performance on unseen target domains; FedSR~\cite{FedSR} enforces an L2-norm regularizer on the representation and conditional mutual information to enable the model to learn a simple representation of the data for better generalization.
Furthermore, non-IID domain feature settings can also be considered as cross-domain FL, i.e. each client represents a specific domain where PartialFed~\cite{Partialfed} performs local optimization of selective model parameters and does not initialize them as a global model. 

\subsection{Disentangled feature learning}
Learning good feature representations is crucial to ensure the performance of computer vision algorithms. Disentangled feature learning, as a simple and effective feature representation method, has been widely explored in computer vision tasks, such as identifying sources of variation for interpretability~\cite{higgins2016beta,jeong2019learning,chen2018isolating}, obtaining representation invariant to nuisance factors~\cite{moyer2018invariant,song2019learning,tishby2000information}, and domain transfer~\cite{liu2022learning,hwang2020variational,press2020emerging}. 
In domain adaptation and generalization, some studies have also improved the performance by introducing Fourier transforms to decouple the feature information in images, and achieved promising performance~\cite{FedDG,xu2021fourier}. 
Currently, most previous FL approaches attempt to improve performance at the model level, few introduce disentangled feature learning into the FL to explore non-IID domain features from in latent feature space, which resulted in insufficient learning of the correlation of global domain-invariant and local domain-specific features, thus leading to a limited performance on FL tasks with large differences in feature distribution between clients.
Therefore, in this paper, we design a novel disentangled feature learning-based FL framework to enable the network to learn the global domain-invariant representation well while preserving the local domain-specific feature learning.

\begin{figure*}[!t]
\vspace{4pt}
 \begin{center}
  \includegraphics[width=1\linewidth]{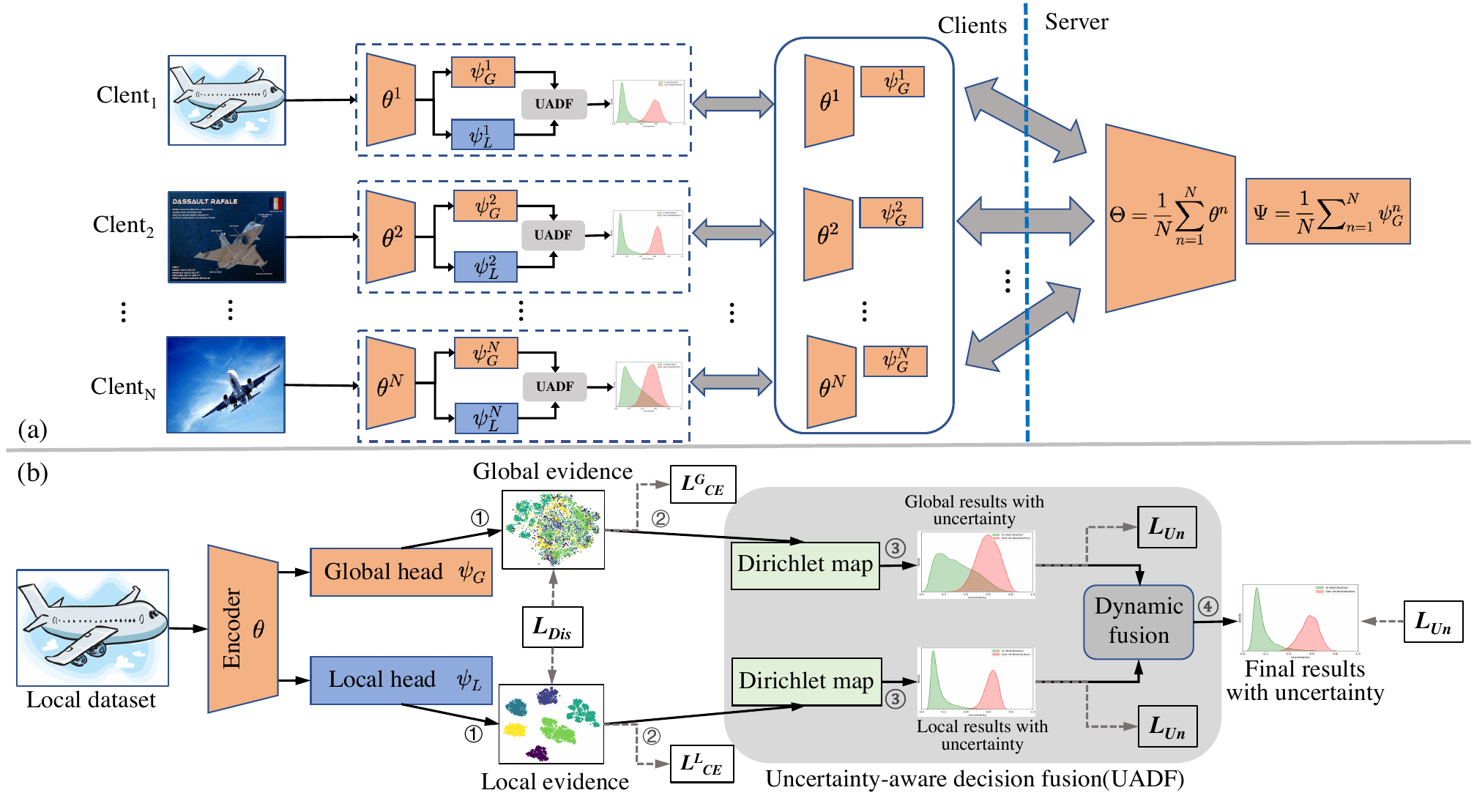}
 \end{center}
 \captionsetup{font=small}
 \vspace{-8pt}
 \caption{\small{The framework of the proposed RFedDis. (a) Overview of the proposed FL collaborative paradigm. (b) The detailed chart of local client training. The $L_{Dis}$ is used to enlarge the feature distribution differences between the global and local heads to guide the feature disentangling, and $L_{Un}$ is adopted to guide the optimization of the prediction based on the feature distribution that was parameterized by Dirichlet concentration, while $L_{CE}^G$ and $L_{CE}^G$ are attached to the decoupled original features of the global and local heads, respectively, to ensure the confidence of the decoupled original features during iterative fusion optimization.}}
 \vspace{-10pt}
 \label{Fig2}
\end{figure*}

\subsection{Uncertainty estimation.}
Uncertainty estimation methods play a pivotal role in evaluating the confidence and trustworthiness of the model during both optimization and decision-making processes. It enables the AI system to make reliable decisions and avoid potential disasters caused by OOD samples~\cite{FirstUncertainty,guynn2015google,nhtsa2016department}.
There are many studies have been proposed. For example, MC-Dropout~\cite{MC_Drop} obtains model uncertainty by casting dropout training in deep neural networks as approximate Bayesian inference in deep Gaussian processes. Lakshminarayanan \textit{et.al.}~proposed a non-Bayesian uncertainty estimation solution using deep ensembles~\cite{EnsembleUncertainty}. Instead of indirectly modeling uncertainty through network weights, Sensoy \textit{et.al.}~model uncertainty directly with subjective logical (SL) evidential theory~\cite{Deterministic}. 
Furthermore, Amersfoort \textit{et.al.}~proposed a deterministic uncertainty quantiﬁcation method based on the idea of modeling uncertainty with RBF network in a deep neural network in a single forward pass~\cite{DeterministicUncertainty}. 
In addition, uncertainty theory has also been introduced to explore the technical challenges in multi-view/model learning. Kendall \textit{et.al.}~achieved impressive performance in multitask learning by obtaining learning weights for different tasks through homoskedasticity uncertainty learning~\cite{MultiTaskUncetainty}. More recently, Han \textit{et.al.}~proposed a trusted multi-view classiﬁcation algorithm based on the Dempster-Shafer evidence theory~\cite{TMC}. However, unlike multi-view tasks with complementary and deterministic domain feature information, FL task with non-IID domain features suffers from severe heterogeneous non-determinism in feature distribution between clients. Such behavior leads to the under-performance of FL tasks with non-IID domain features. Therefore, we introduce Uncertainty-aware decision fusion combined with our improved objective function to dynamically integrate decoupled features at the evidence level, while generating reliable predictions with estimated uncertainty without loss of accuracy.

\section{Proposed Method}
\label{sec:Trusted}

In this section, we present the design of our RFedDis framework, as shown in Fig.~\ref{Fig2}, which applies the federated disentangling to decouple the global domain-invariant feature and  client-specific feature, and utilizes an Uncertainty-aware decision fusion for dynamically integrating the decoupled features at the evidence level, and producing a reliable prediction with an estimated uncertainty.

\subsection{Federated feature disentangling}

\begin{figure*}[!t]
\vspace{4pt}
 \begin{center}
  \includegraphics[width=0.8\linewidth,height=0.6\linewidth]{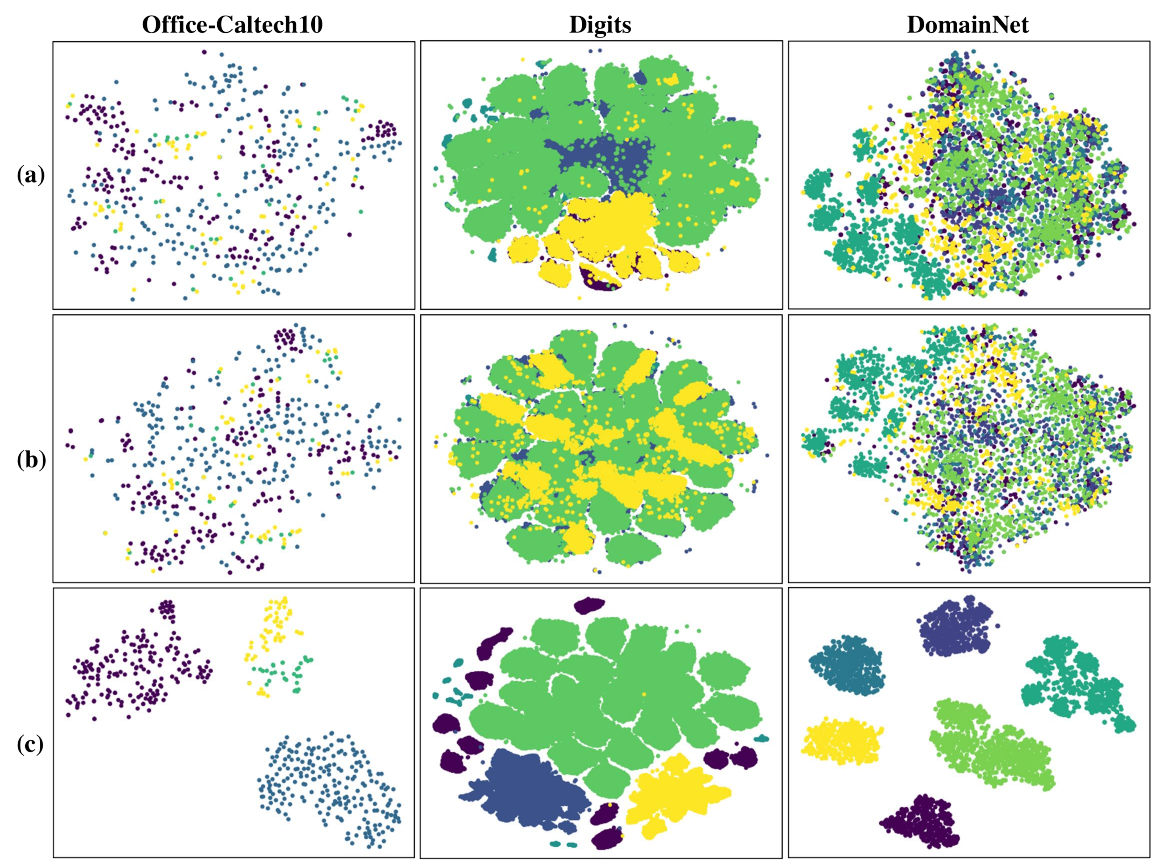}
 \end{center}
 \vspace{-9pt}
 \caption{\small{T-SNE visualization for the hidden features of our proposed network on datasets of OfficeCaltech10, Digits, and DomainNet, where each color represents a client. Compared to the encoder features (a), the global head features (b) have a more general feature distribution and the inter-domain feature gap is mitigated, while the local head features (c) mainly preserve the client-specific information.}}
 \label{T-SNE}
 \vspace{-10pt}
\end{figure*}

In FL setting, non-IID domain feature usually means that the feature distributions $P\left(X \right)$ are different between clients but with the same marginal distributions of the labels $P\left(Y \right)$. A typical example in practice is that data collected by different clients may come from different acquisition devices with different styles or contrasts, \textit{i.e.}, $P_{i}\left(Y\right)=P_{j}\left( Y\right)$, and  $P_{i}\left( Y|X\in D_{i}\right)\neq P_{j}\left( Y|X\in D_{j}\right)$. Therefore, given the input data with multiple domain features, how to efficiently learn a global generic domain-invariant representation while preserving local client-specific properties is crucial for FL on non-IID data. Most existing FL approaches mainly focus on learning global cross-client models and ignore the local client-specific properties~\cite{FedAvg,FedBN,FedDC}. 
Recently, some previous studies have found that the shallow layers of the network prefer learning global generic features, and as the network goes deeper, the network gradually focuses on learning category-related features~\cite{CKA,9880164}. 
Meanwhile, we visualize the features of the proposed model from different clients by using T-SNE, as shown in Fig.~\ref{T-SNE}.
The encoder of FL model often captures the global generic cross-client information (Fig.~\ref{T-SNE} (a)), but the features between different domains in the learned feature space are not well aligned. 
The inter-domain feature differences can still be observed, which may lead to interference between domain-specific feature and cross-client generic information resulting in sub-optimal performance. 
Therefore, decoupling the general domain-invariant information and local domain-specific features from the shallow encoder layers is crucial to deal with the non-IID domain feature. 
Importantly, it can enhance the model's capacity of learning local domain-specific features while alleviating the interference of local domain-specific features with the global generic cross-client information.

Therefore, we design a novel feature disentangling FL framework that consists of three main components: encoder path $( \theta^{1},\theta^{2},...,\theta^{N} )  \in \Theta$, global head $( \psi^{1}_{G} ,\psi^{2}_{G} ,...,\psi^{N}_{G} )\in\Psi_{G}$, and local head $( \psi^{1}_{L},\psi^{2}_{L} ,...,\psi^{N}_{L} )\in\Psi_{L}$, where $N$ is the total number of clients, and $\theta_{i}, \psi^{i}_{G}$, and $\psi^{i}_{L}$ represent the model parameters of the encoder, global head branch, and local head path of \textit{i}-th client, respectively. As shown in Fig.~\ref{Fig2}, the encoder is used to capture the global generic cross-client information through collaborative FL learning, while the global and local heads are adopted to further decouple the encoder feature into the generic domain-invariant cross-client representation and local domain-specific features, as:
\begin{equation}
F^{i}_{G}=\psi^{i}_{G} \left( \theta^{i}\left( X\right)  \right),  
F^{i}_{L}=\psi^{i}_{L} \left( \theta^{i} \left( X\right)  \right),  
\label{eq:1}
\end{equation}
where $F^{i}_{G}$ and $F^{i}_{L}$ are the global domain-invariant cross-client information and local domain-specific features learned by the $\textit{i}$-th client. As shown in Fig.~\ref{Fig2}, to maintain the capacity of the global head to represent the global generic cross-client information while alleviating the interference of domain-specific features, the global head is shared with the server for model aggregation while the local head is updated in client for personalized learning. Moreover, an inverse KL divergence ($L_{Dis}$) is introduced to maximize the feature distribution differences between the global head and the local head, as:
\begin{equation}
L_{Dis}=exp^{-D_{KL}\left( P\| Q\right)},
\label{InverseKL}
\end{equation}
with
\begin{equation}
\setlength{\abovedisplayskip}{3pt}
\setlength{\belowdisplayskip}{3pt}
D_{KL}\left( P\| Q\right)  =\sum P\left( F_{L};X\right)  \log \dfrac{P\left(F_{L};X\right)}{Q\left( F_{G};X\right)},
\label{eq:3}
\end{equation}
where $P\left(F_{L};X\right)$ and $Q\left(F_{G};X\right)$ are the feature distributions of the global head and the local head by given input \textit{X}. 
It can be seen from Eq.~\ref{InverseKL} that $L_{Dis}$ inversely proportional to the KL distance between $P\left( F_{L};X\right)$ and $Q\left( F_{G};X\right)$, \textit{i.e.}, if the feature information of $P\left( F_{L};X\right)$ and $Q\left( F_{G};X\right)$ are close to each other, the KL is small and thus generates a larger $L_{dis}$, and conversely, if $P\left( F_{L};X\right)$ and $Q\left( F_{G};X\right)$ with a large difference, the KL increases and thus generates a smaller $L_{Dis}$. 
Therefore, during the training process, the feature distance between $F_{G}$ and $F_{L}$ can be enlarged by minimizing Eq.~\ref{InverseKL} to achieve the decoupling of $F_{G}$ and $F_{L}$ feature distribution.

With the federated disentangling, we could decouple the global domain-invariant feature and client-specific feature. Fig.~\ref{T-SNE} (b) and (c) show T-SNE feature visualization of the global head and local head in our model.
Compared to the aggregated encoder feature (Fig.~\ref{T-SNE} (a)), the global head features have a more general feature distribution, and the inter-domain feature gap is also mitigated. Instead, the local head preserves the domain-specific features from the encoder features by maximizing the feature distribution differences between the local head and the global head. It mainly keeps the client-specific domain information (Fig.~\ref{T-SNE} (c)), compared to the aggregated encoder and global head. 

\subsection{Uncertainty-aware decision fusion}
How to effectively fuse the decoupled features and generate a reliable decision with an estimated uncertainty is another challenge to be explored in this paper. 
To this end, we introduce Uncertainty-aware decision fusion  based on evidence-based subjective logic (SL) uncertainty\cite{Deterministic} to derive the prediction distributions and corresponding overall uncertainty score for the global and local heads, respectively.
Then, the prediction results of both heads are dynamically fused to obtain the final decision with corresponding uncertainty evaluation. Specifically, suppose that there is a $K$ categories problem, the $K+1$ mass values ($K$ belief masses and one corresponding uncertainty mass) are all non-negative and their sum is one, \textit{i.e.}, $u+\sum^{K}_{i=1} b_{i}=1$, where $b_{k}\geq0$ and $u\geq0$ are the probability for the \textit{k}-th category and the overall uncertainty score, respectively. 
Therefore, as shown in Fig.~\ref{Fig2} (b), the belief masses and overall uncertainty scores can be calculated by the following four steps: 

\noindent\textbf{Step \circled{1}:} Obtaining the evidence $E_{G} = [e^{k}_{G}]$  and $E_{L} = [e^{k}_{L}]$ for the global and the local heads, respectively, by applying \textit{Softplus} activation function to ensure the feature values are larger than 0:
\begin{equation}
\begin{split}
E_{G} &=Softplus\left( \psi_{G} \left( \theta \left( X\right)  \right)  \right),\\ 
E_{L} &=Softplus\left( \psi_{L} \left( \theta \left( X\right)  \right)  \right).
\end{split}
  \label{eq:4}
\end{equation}

\noindent\textbf{Step \circled{2}:} Parameterizing  $E_{G}$ and $E_{L}$ to Dirichlet distribution, as:
\begin{equation} 
\begin{split}
\alpha_{G} =& E_{G}+1, \; i.e., \; \alpha^{k}_{G} =e^{k}_{G}+1,\\
\alpha_{L} =& E_{L}+1, \; i.e., \; \alpha^{k}_{L} =e^{k}_{L}+1,
\end{split}
\label{eq:5}
\end{equation}
where $\alpha^{k}_{G}$ and $e^{k}_{G}$ are the \textit{k}-th category Dirichlet distribution parameters and evidence of global head, while $\alpha^{k}_{L}$ and $e^{k}_{L}$ indicate the \textit{k}-th category Dirichlet distribution parameters and evidence of local head, respectively.

\noindent\textbf{Step \circled{3}:} Calculating the belief masses and corresponding uncertainty scores for each head as:
\begin{equation}
\begin{split}
b^{k}_{G}=\frac{e^{k}_{G}}{S_{G}} =\frac{\alpha^{k}_{G} -1}{S_{G}} ,  \;  u_{G}=\frac{K}{S_{G}},\\
b^{k}_{L}=\frac{e^{k}_{L}}{S_{L}} =\frac{\alpha^{k}_{L} -1}{S_{L}}, \; u_{L}=\frac{K}{S_{L}},
\end{split}
\label{eq:6}
\end{equation}
where $S_{G}=\sum\nolimits^{K}_{k=1} \left( e^{k}_{G}+1\right) =\sum\nolimits^{K}_{k=1} \alpha^{k}_{G}$ and $S_{L}=\sum\nolimits^{K}_{k=1} \left( e^{k}_{L}+1\right) =\sum\nolimits^{K}_{k=1} \alpha^{k}_{L}$ are the Dirichlet intensities of the global head and local head features, respectively. Therefore, it can be seen from Eq.~(\ref{eq:6}), the probability assigned to category \textit{k} is proportional to the observed evidence for category \textit{k}. Conversely, if less total evidence is obtained, the greater the total uncertainty.

\noindent \textbf{Dynamic decision fusion (Step \circled{4})} 
Finally, we apply a dynamic decision fusion to combine the belief masses and overall uncertainty scores of these two heads. The joint mass $\textit{M}$=$\left\{b_{1},b_{2},...,b_{K},\  u\right\}$ can be formulated based on Dempster-Shafer~\cite{TMC} theory:
\begin{equation} 
M=M_{G}\oplus M_{L},
\label{eq:7}
\end{equation}
\noindent where $M_{G}\!=\!\left\{ b^{1}_{G},b^{2}_{G},...,b^{K}_{G},u_{G}\right\}$ and $M_{L}=\left\{b^{1}_{L},b^{2}_{L},...,b^{K}_{L},u_{L}\right\}$ are the masses of global and local heads. Specifically,
\begin{equation}
\begin{aligned}
b_{k}&=\dfrac{\left( b^{k}_{G}b^{k}_{L}+b^{k}_{G}u_{L}+b^{k}_{L}u_{G}\right)}{1-C} , \; \\ 
\;\!u&=\dfrac{u_{G}u_{L}}{1-C} ,  
\label{eq:8}
\end{aligned}
\end{equation}
where $C=\sum\nolimits_{i\neq j} b^{i}_{G}b^{j}_{L}$ denotes the amount of conflict between two mass sets. Furthermore, it can be seen intuitively from Eq.~(\ref{eq:8}) that 1) If both heads have high uncertainty, \textit{i.e.}, large $u_{G}$ and $u_{L}$, then the final prediction will be of low confidence, \textit{i.e.}, small $b_{k}$. 2) Conversely, if both heads have low uncertainty, \textit{i.e.}, small $u_{G}$ and $u_{L}$, then the final prediction will be of high confidence, \textit{i.e.}, large $b_{k}$. 3) If only one head is low uncertainty, \textit{i.e.}, only $u_{G}$ or $u_{L}$ is large, then the final prediction depends on the more confident head. Therefore, based on dynamic decision fusion, the FL model can further improve the reliability of decisions and jointly support the reliable prediction. 

\subsection{Loss function}
As shown in Fig.~\ref{Fig2} (b), our proposed RFedDis mainly contains two components: federated disentangling and Uncertainty-aware decision fusion. 

\noindent \textbf{In federated disentangling,} we use $L_{Dis}$ in Eq.~(\ref{InverseKL}) to enlarge the feature distribution differences between the global and local heads. Moreover, we also introduce the cross-entropy loss on the decoupled original features of the global and local heads (\textit{i.e.}, $L_{CE}^L$ and $L_{CE}^G$) to directly guide the decoupled feature distribution.
The cross-entropy loss ($L_{CE}$) is the commonly used loss function for optimizing neural network-based classifiers,
\begin{equation}
\setlength{\abovedisplayskip}{3pt}
\setlength{\belowdisplayskip}{3pt}
L_{CE}=-\sum^{K}_{k=1} y_{m,k}log\left(b_{m,k}\right),  
  \label{eq:10}
\end{equation}
where $y_{m,k}$ and $b_{m,k}$ are the label and probability on the \textit{k}-th category of \textit{m}-th sample, respectively. 

\noindent \textbf{In Uncertainty-aware decision fusion,} we introduce the uncertainty loss $L_{Un}$ to guide the optimization of the prediction based on the feature distribution that was parameterized by Dirichlet concentration. Similar to previous uncertainty works~\cite{TMC} that directly obtains deterministic predictions given input features, in our proposed RFedDis, SL is employed to associate the evidence with the parameters of Dirichlet distribution, i.e, given the evidence of $E=\left\{ e_{1},e_{2},...,e_{K}\right\}$, we can get the Dirichlet distribution parameter of $\alpha=E+1$ and the category belief mass of $\emph{b}=\left\{b_{1},b_{2},...,b_{K}|\alpha \right\}$. Therefore, based on the cross-entropy loss in Eq.~(\ref{eq:10}), the objective function for optimizing the re-parameterized evidence can be further formulated as follows:
\begin{equation} 
\footnotesize
\begin{split}
L_{Dce}\! &=\!\int \left[ \sum^{K}_{k=1} -y_{m,k} \log \left( b_{m,k}\right)  \right]  \frac{1}{\beta \left( \alpha_{m} \right)} \prod^{K}_{k=1} b^{\alpha_{m,k} -1}_{m,k}db_{m}  \\
&=\sum^{K}_{k=1} y_{m,k}\left( \Phi \left( S_{m}\right)  -\Phi \left( \alpha_{m,k} \right)  \right), 
\end{split}
\label{eq:11}
\end{equation}
where $\Phi (\cdot)$ denotes the digamma function, while $\beta \left( \alpha_{m} \right)$ is the multinomial beta function for the \textit{m}-th sample concentration parameter $\alpha_{m}$, and $K$ indicates total number of categories. Meanwhile, we further introduce the KL divergence function to ensure that incorrect labels will yield less evidence: 
\begin{equation}
\small
\begin{split}
L_{KL} &= \log \left( \dfrac{\Gamma \left( \sum^{K}_{k=1} \left( \tilde{\alpha }_{m,k} \right) \right)  }{\Gamma \left( K\right)  \sum^{K}_{k=1} \Gamma \left( \tilde{\alpha }_{m,i} \right)} \right) \\ 
&+\sum_{k=1}^{K} \left( \tilde{\alpha }_{m,k} -1\right)  \left[ \Phi \left( \tilde{\alpha }_{m,k} \right)  -\Phi \left( \sum_{i=1}^{K} \tilde{\alpha }_{m,k} \right)  \right], 
\end{split}
\label{eq:11-L}
\end{equation} 
where $\Gamma (\cdot)$ is the gamma function, while $\tilde{\alpha }_{m} =y_{m}+\left( 1-y_{m}\right) \odot \alpha_{m}$ denotes the adjusted parameters of the Dirichlet distribution which aims to avoid penalizing the evidence of the ground-truth class to 0. Therefore, the objective function for the model optimization based on the feature distribution parameterized by Dirichlet concentration is as follows:
\begin{equation}
L_{Un}=L_{Dce}+\lambda_{u} \ast L_{KL}, 
\label{eq:12}
\end{equation}
where $\lambda_{u}$ is the balance factor weighted to $L_{KL}$. To prevent the network from focusing too much on KL divergence in the initial phase of training, which may lead to a lack of good exploration of the parameter space and causes the network to output a flat uniform distribution, we initialize $\lambda_{u}$ as 0 and increase it gradually with the number of training iterations.
The uncertainty loss $L_{Un}$ will work on the global head, local head, and fused decision fusion, respectively, as shown in  Fig.~\ref{Fig2} (b).

Note that uncertainty loss $L_{Un}$ is adopted to guide the optimization of the prediction based on the feature distribution which was parameterized by Dirichlet concentration. This will change the original decoupled feature distribution of the global and local heads, which may cause a decline in the classifier’s confidence for the decoupled features and affect the feature disentangling optimization, thus resulting in limited performance on non-IID domain feature. 
Therefore, to ensure the confidence of the decoupled original features during the iterative fusion optimization, we apply cross-entropy loss ($L_{CE}^L$ and $L_{CE}^G$) on the decoupled features of the global and local heads to
directly guide the decoupled feature distribution.
Overall, the objective function is:
\begin{equation}
L_{all}=L_{Un}+L_{CE}^L +L_{CE}^G +\lambda_{d} \ast L_{Dis} ,
  \label{Lall}
\end{equation}
where $L_{Dis}$ is used to enlarge the feature distribution differences between the global and local heads by Eq.~(\ref{InverseKL}), and $\lambda_{d}$ is a balancing factor on $L_{Dis}$. To prevent the network from paying too much attention to the distribution difference between $F_{G}$ and $F_{L}$ during the initial training phase, we initialize $\lambda_{d}$ as 0 and
increase it gradually with the number of training iterations.  
\section{Experiments}
\label{Experiments}
\subsection{Experimental setup}
In this paper, we are focusing on the non-IID domain feature problem, following the similar setting used in FedBN~\cite{FedBN}, \textit{e.g.}, an SGD optimizer with a learning rate of 1e-2 to optimize the model, a batch size set to 32, and the same data pre-processing strategy. We also perform the 5 trials of repeating experiments with different random seeds and report the mean and variance of the test accuracy for each dataset. In addition, except for the introduction of three fully connected (FC) layers as the local head, we use the same architecture as FedBN, \textit{i.e.}, three convolutional layers connected to three FC layers.

\textbf{Dataset:} Since the medical dataset used in FedBN is not publically available according to its GitHub page, we conduct extensive experiments on three other natural image datasets: 1$)$ \textbf{OfficeCaltech10}~\cite{OfﬁceCaltech10}, which has four data sources, namely Office-3 (contains three data sources: amazon (A), dslr (D) and webcam (W)), and Caltech-256 (C) datasets (captured by different camera devices or in different real-world environments with various contexts). 2$)$ \textbf{Digits}, which consists of the following five datasets from different domains: MNIST (M)~\cite{Digits_MNIST}, SVHN (S)~\cite{Digits_SVHN}, USPS(U)~\cite{Digits_USPS}, SynthDigits (Syn)~\cite{Digits_MNISTM}, and MNIST-M (MM)~\cite{Digits_MNISTM}. 3$)$ \textbf{DomainNet}~\cite{DomainNet}, which contains natural images from six different data sources with different image styles: Clipart (Cl), Infograph (I), Painting (P), Quickdraw (Q), Real (R), and Sketch (Sk). 

\textbf{Baselines:} To comprehensively evaluate the performance of our proposed RFedDis, in addition to commonly used baselines FedAvg~\cite{FedAvg} and FedProx~\cite{FedProx}, we also analyze several SOTA FL methods personalized for non-IID tasks, including FedRep~\cite{FedRep}, FedDC~\cite{FedDC}, FedDyn~\cite{FedDyn}, Moon~\cite{Moon}, SCAFFOLD~\cite{SCAFFOLD}, and FedBN~\cite{FedBN}.
\subsection{Comparison results}
Table~\ref{tab:Office-Caltech10} and Table~\ref{tab:DigitsAndDomainNet} show the test accuracy performance of the different approaches on Office-Caltech10, Digits, and DomainNet, respectively. From Table~\ref{tab:Office-Caltech10} and Table~\ref{tab:DigitsAndDomainNet}, we can summarize the following observations:

1) The proposed RFedDis achieves the highest test accuracy on most datasets, which indicates that our proposed RFedDis is more effective than other FL methods in solving non-IID domain features. On Office-Caltech10, our proposed RFedDis achieves the highest test accuracy on all clients, and the average accuracy of the proposed method is improved by 7.5\% compared to FedBN. Meanwhile, as shown in Table~\ref{tab:DigitsAndDomainNet}, our proposed RFedDis achieves a 1.7\% and 2.5\% improvement over the suboptimal baseline methods FedDyn~\cite{FedDyn} and FedRep~\cite{FedRep} on Digits and DomainNet, respectively.
\begin{table}[!t]
\footnotesize 
  \centering
 \captionsetup{font=small}
  \vspace{4pt}
 \caption{\small{The detailed statistics reported with format mean (std) of test accuracy among different approaches on Office-Caltech10.}}
  \vspace{-4pt}
\resizebox{1\linewidth}{!}{
\begin{tabular}{l|cccc|c}
\toprule[1pt]
Methods & \multicolumn{1}{c|}{A} & \multicolumn{1}{c|}{C} & \multicolumn{1}{c|}{D} & \multicolumn{1}{c|}{W} & Avg \\ \bottomrule[1pt]
SingleSet & \begin{tabular}[c]{@{}c@{}}54.9 (1.5)\end{tabular} & \begin{tabular}[c]{@{}c@{}}40.2(1.6)\end{tabular} & \begin{tabular}[c]{@{}c@{}}78.7 (1.3)\end{tabular} & \begin{tabular}[c]{@{}c@{}}86.4(2.4)\end{tabular} & \begin{tabular}[c]{@{}c@{}}65.1(1.7)\end{tabular} \\ \hline
FedAvg\cite{FedAvg} & \begin{tabular}[c]{@{}c@{}}54.1(1.1)\end{tabular} & \begin{tabular}[c]{@{}c@{}}44.8(1.0)\end{tabular} & \begin{tabular}[c]{@{}c@{}}66.9(1.5)\end{tabular} & \begin{tabular}[c]{@{}c@{}}85.1(2.9)\end{tabular} & \begin{tabular}[c]{@{}c@{}}62.7(1.6)\end{tabular} \\ \hline
FedProx\cite{FedProx} & \begin{tabular}[c]{@{}c@{}}54.2(2.5)\end{tabular} & \begin{tabular}[c]{@{}c@{}}44.5(0.5)\end{tabular} & \begin{tabular}[c]{@{}c@{}}65.0(3.6)\end{tabular} & \begin{tabular}[c]{@{}c@{}}84.4(1.7)\end{tabular} & \begin{tabular}[c]{@{}c@{}}62.0(2.1)\end{tabular} \\ \hline
FedRep\cite{FedRep} & \begin{tabular}[c]{@{}c@{}}51.0(1.7)\end{tabular} & \begin{tabular}[c]{@{}c@{}}40.9(0.8)\end{tabular} & \begin{tabular}[c]{@{}c@{}}62.5(1.8)\end{tabular} & \begin{tabular}[c]{@{}c@{}}79.7(2.1)\end{tabular} & \begin{tabular}[c]{@{}c@{}}58.5(1.6)\end{tabular} \\ \hline
Moon\cite{Moon} & \begin{tabular}[c]{@{}c@{}}55.2(2.3)\end{tabular} & \begin{tabular}[c]{@{}c@{}}46.7(1.1)\end{tabular} & \begin{tabular}[c]{@{}c@{}}68.8(2.6)\end{tabular} & \begin{tabular}[c]{@{}c@{}}84.8(2.1)\end{tabular} & \begin{tabular}[c]{@{}c@{}}63.9(2.0)\end{tabular} \\ \hline
FedDC\cite{FedDC} & \begin{tabular}[c]{@{}c@{}}61.5(1.8)\end{tabular} & \begin{tabular}[c]{@{}c@{}}40.9(1.2)\end{tabular} & \begin{tabular}[c]{@{}c@{}}81.3(2.2)\end{tabular} & \begin{tabular}[c]{@{}c@{}}81.4(1.8)\end{tabular} & \begin{tabular}[c]{@{}c@{}}66.3(1.8)\end{tabular} \\ \hline
FedDyn\cite{FedDyn} & \begin{tabular}[c]{@{}c@{}}55.7(1.1)\end{tabular} & \begin{tabular}[c]{@{}c@{}}40.0(0.7)\end{tabular} & \begin{tabular}[c]{@{}c@{}}81.3(1.9)\end{tabular} & \begin{tabular}[c]{@{}c@{}}79.7(2.1)\end{tabular} & \begin{tabular}[c]{@{}c@{}}64.2(1.5)\end{tabular} \\ \hline
SCAFFOLD\cite{SCAFFOLD} & \begin{tabular}[c]{@{}c@{}}60.4(1.1)\end{tabular} & \begin{tabular}[c]{@{}c@{}}45.8(1.3)\end{tabular} & \begin{tabular}[c]{@{}c@{}}84.4(2.3)\end{tabular} & \begin{tabular}[c]{@{}c@{}}88.1(2.1)\end{tabular} & \begin{tabular}[c]{@{}c@{}}69.7(1.7)\end{tabular} \\ \hline
FedBN\cite{FedBN} & \begin{tabular}[c]{@{}c@{}}63.0(1.6)\end{tabular} & \begin{tabular}[c]{@{}c@{}}45.3(1.5)\end{tabular} & \begin{tabular}[c]{@{}c@{}}83.1(2.5)\end{tabular} & \begin{tabular}[c]{@{}c@{}}90.5(2.3)\end{tabular} & \begin{tabular}[c]{@{}c@{}}70.5(2.0)\end{tabular} \\ \hline
\textbf{\begin{tabular}[c]{@{}c@{}}Our RFedDis\end{tabular}} & \textbf{\begin{tabular}[c]{@{}c@{}}64.5(1.0)\end{tabular}} & \textbf{\begin{tabular}[c]{@{}c@{}}48.2(0.9)\end{tabular}} & \textbf{\begin{tabular}[c]{@{}c@{}}95.3(3.1)\end{tabular}} & \textbf{\begin{tabular}[c]{@{}c@{}}95.3(2.1)\end{tabular}} & \textbf{\begin{tabular}[c]{@{}c@{}}75.8(1.8)\end{tabular}} \\ 
\bottomrule[1pt]
\end{tabular}
\label{tab:Office-Caltech10}
}
\vspace{-10pt}
\end{table}

\begin{table*}[!t]
\vspace{8pt}
 \captionsetup{font=small}
 \caption{\small{The detailed statistics reported with format mean (std) of test accuracy among different approaches on Digits and DomainNet.}}
 \vspace{-3pt}
  \centering
\resizebox{1.\textwidth}{!}{
\begin{tabular}{l|ccccc|c||cccccc|c}
\toprule[1pt]
 & \multicolumn{6}{c||}{Digits} & \multicolumn{7}{c}{DomainNet} \\ \cline{2-14} 
\multirow{-2}{*}{Methods} & \multicolumn{1}{c|}{M} & \multicolumn{1}{c|}{S} & \multicolumn{1}{c|}{U} & \multicolumn{1}{c|}{Syn} & \multicolumn{1}{c|}{MM} & Avg & \multicolumn{1}{c|}{Cl} & \multicolumn{1}{c|}{I} & \multicolumn{1}{c|}{P} & \multicolumn{1}{c|}{Q} & \multicolumn{1}{c|}{R} & \multicolumn{1}{c|}{Sk} & \multicolumn{1}{c}{Avg} \\ \bottomrule[1pt]
SingleSet & \begin{tabular}[c]{@{}c@{}}94.4(0.1)\end{tabular} & \begin{tabular}[c]{@{}c@{}}65.3(1.1)\end{tabular} & \begin{tabular}[c]{@{}c@{}}95.2(0.1)\end{tabular} & \begin{tabular}[c]{@{}c@{}}80.3(0.4)\end{tabular} & \begin{tabular}[c]{@{}c@{}}77.8(0.5)\end{tabular} & \begin{tabular}[c]{@{}c@{}}82.0(0.4)\end{tabular} & \begin{tabular}[c]{@{}c@{}}41.0(0.9)\end{tabular} & \begin{tabular}[c]{@{}c@{}}23.8(1.2)\end{tabular} & \begin{tabular}[c]{@{}c@{}}36.2(2.7)\end{tabular} & \begin{tabular}[c]{@{}c@{}}73.1(0.9)\end{tabular} & \begin{tabular}[c]{@{}c@{}}48.5(1.9)\end{tabular} & \begin{tabular}[c]{@{}c@{}}34.0(1.1)\end{tabular} & \begin{tabular}[c]{@{}c@{}}42.8(1.5)\end{tabular} \\ \hline
FedAvg\cite{FedAvg} & \begin{tabular}[c]{@{}c@{}}95.9(0.2)\end{tabular} & \begin{tabular}[c]{@{}c@{}}62.9(1.5)\end{tabular} & \begin{tabular}[c]{@{}c@{}}95.6(0.3)\end{tabular} & \begin{tabular}[c]{@{}c@{}}82.3(0.4)\end{tabular} & \begin{tabular}[c]{@{}c@{}}76.9(0.5)\end{tabular} & \begin{tabular}[c]{@{}c@{}}82.7(0.6)\end{tabular} & \begin{tabular}[c]{@{}c@{}}48.8(1.9)\end{tabular} & \begin{tabular}[c]{@{}c@{}}24.9(0.7)\end{tabular} & \begin{tabular}[c]{@{}c@{}}36.5(1.1)\end{tabular} & \begin{tabular}[c]{@{}c@{}}56.1(1.6)\end{tabular} & \begin{tabular}[c]{@{}c@{}}46.3(1.4)\end{tabular} & \begin{tabular}[c]{@{}c@{}}36.6(2.5)\end{tabular} & \begin{tabular}[c]{@{}c@{}}41.5(1.5)\end{tabular} \\ \hline
FedProx\cite{FedProx} & \begin{tabular}[c]{@{}c@{}}95.8(0.2)\end{tabular} & \begin{tabular}[c]{@{}c@{}}63.1(1.6)\end{tabular} & \begin{tabular}[c]{@{}c@{}}95.6(0.3)\end{tabular} & \begin{tabular}[c]{@{}c@{}}82.3(0.4)\end{tabular} & \begin{tabular}[c]{@{}c@{}}76.6(0.6)\end{tabular} & \begin{tabular}[c]{@{}c@{}}82.7(0.6)\end{tabular} & \begin{tabular}[c]{@{}c@{}}48.9(0.8)\end{tabular} & \begin{tabular}[c]{@{}c@{}}24.9(1.0)\end{tabular} & \begin{tabular}[c]{@{}c@{}}36.6(1.8)\end{tabular} & \begin{tabular}[c]{@{}c@{}}54.4(3.1)\end{tabular} & \begin{tabular}[c]{@{}c@{}}47.8(0.8)\end{tabular} & \begin{tabular}[c]{@{}c@{}}36.9(2.1)\end{tabular} & \begin{tabular}[c]{@{}c@{}}41.6(1.6)\end{tabular} \\ \hline
FedRep\cite{FedRep} & \begin{tabular}[c]{@{}c@{}}96.8(0.2)\end{tabular} & \begin{tabular}[c]{@{}c@{}}71.2(1.7)\end{tabular} & \begin{tabular}[c]{@{}c@{}}96.9(0.2)\end{tabular} & \begin{tabular}[c]{@{}c@{}}82.7(0.4)\end{tabular} & \begin{tabular}[c]{@{}c@{}}78.6(0.7)\end{tabular} & \begin{tabular}[c]{@{}c@{}}85.2(0.6)\end{tabular} & \begin{tabular}[c]{@{}c@{}}52.5(1.2)\end{tabular} & \begin{tabular}[c]{@{}c@{}}27.3(1.1)\end{tabular} & \begin{tabular}[c]{@{}c@{}}41.5(1.8)\end{tabular} & \begin{tabular}[c]{@{}c@{}}70.8(2.6)\end{tabular} & \begin{tabular}[c]{@{}c@{}}55.9(2.3)\end{tabular} & \textbf{\begin{tabular}[c]{@{}c@{}}43.7(1.9)\end{tabular}} & \begin{tabular}[c]{@{}c@{}}48.6(1.8)\end{tabular} \\ \hline
Moon\cite{Moon} & \begin{tabular}[c]{@{}c@{}}95.7(0.3)\end{tabular} & \begin{tabular}[c]{@{}c@{}}62.7(1.8)\end{tabular} & \begin{tabular}[c]{@{}c@{}}95.3(0.2)\end{tabular} & \begin{tabular}[c]{@{}c@{}}80.8(0.5)\end{tabular} & \begin{tabular}[c]{@{}c@{}}75.5(0.6)\end{tabular} & \begin{tabular}[c]{@{}c@{}}82.0(0.7)\end{tabular} & \begin{tabular}[c]{@{}c@{}}47.7(0.8)\end{tabular} & \begin{tabular}[c]{@{}c@{}}25.4(0.7)\end{tabular} & \begin{tabular}[c]{@{}c@{}}38.6(1.2)\end{tabular} & \begin{tabular}[c]{@{}c@{}}52.6(2.3)\end{tabular} & \begin{tabular}[c]{@{}c@{}}46.7(2.1)\end{tabular} & \begin{tabular}[c]{@{}c@{}}36.3(1.7)\end{tabular} & \begin{tabular}[c]{@{}c@{}}41.2(1.5)\end{tabular} \\ \hline
FedDC\cite{FedDC} & \begin{tabular}[c]{@{}c@{}}96.0(0.3)\end{tabular} & \begin{tabular}[c]{@{}c@{}}64.5(1.6)\end{tabular} & \begin{tabular}[c]{@{}c@{}}94.1(0.3)\end{tabular} & \begin{tabular}[c]{@{}c@{}}81.4(0.4)\end{tabular} & \begin{tabular}[c]{@{}c@{}}73.6(0.7)\end{tabular} & \begin{tabular}[c]{@{}c@{}}81.9(0.7)\end{tabular} & \begin{tabular}[c]{@{}c@{}}46.2(0.8)\end{tabular} & \textbf{\begin{tabular}[c]{@{}c@{}}28.5(1.2)\end{tabular}} & \begin{tabular}[c]{@{}c@{}}37.8(1.1)\end{tabular} & \begin{tabular}[c]{@{}c@{}}60.9(2.3)\end{tabular} & \begin{tabular}[c]{@{}c@{}}51.4(1.7)\end{tabular} & \begin{tabular}[c]{@{}c@{}}30.7(1.3)\end{tabular} & \begin{tabular}[c]{@{}c@{}}42.6(1.4)\end{tabular} \\ \hline
FedDyn\cite{FedDyn} & \begin{tabular}[c]{@{}c@{}}97.3(0.2)\end{tabular} & \begin{tabular}[c]{@{}c@{}}70.7(1.8)\end{tabular} & \begin{tabular}[c]{@{}c@{}}96.8(0.2)\end{tabular} & \begin{tabular}[c]{@{}c@{}}86.4(0.7)\end{tabular} & \begin{tabular}[c]{@{}c@{}}81.4(0.6)\end{tabular} & \begin{tabular}[c]{@{}c@{}}86.5(0.7)\end{tabular} & \begin{tabular}[c]{@{}c@{}}43.8(0.7)\end{tabular} & {\color[HTML]{333333} \begin{tabular}[c]{@{}c@{}}27.6(0.9)\end{tabular}} & \begin{tabular}[c]{@{}c@{}}41.9(1.3)\end{tabular} & \begin{tabular}[c]{@{}c@{}}61.0(2.5)\end{tabular} & \begin{tabular}[c]{@{}c@{}}52.4(1.8)\end{tabular} & \begin{tabular}[c]{@{}c@{}}43.8(2.1)\end{tabular} & \begin{tabular}[c]{@{}c@{}}45.1(1.6)\end{tabular} \\ \hline
SCAFFOLD\cite{SCAFFOLD} & \begin{tabular}[c]{@{}c@{}}97.2(0.3)\end{tabular} & \begin{tabular}[c]{@{}c@{}}72.8(1.6)\end{tabular} & \textbf{\begin{tabular}[c]{@{}c@{}}97.0(0.3)\end{tabular}} & \begin{tabular}[c]{@{}c@{}}84.9(0.5)\end{tabular} & \begin{tabular}[c]{@{}c@{}}80.8(0.7)\end{tabular} & \begin{tabular}[c]{@{}c@{}}86.5(0.7)\end{tabular} & \begin{tabular}[c]{@{}c@{}}48.1(0.6)\end{tabular} & {\color[HTML]{333333} \begin{tabular}[c]{@{}c@{}}27.9(1.1)\end{tabular}} & \begin{tabular}[c]{@{}c@{}}35.5(0.8)\end{tabular} & \begin{tabular}[c]{@{}c@{}}68.5(2.3)\end{tabular} & \begin{tabular}[c]{@{}c@{}}46.7(1.7)\end{tabular} & \begin{tabular}[c]{@{}c@{}}35.4(1.2)\end{tabular} & \begin{tabular}[c]{@{}c@{}}43.7(1.3)\end{tabular} \\ \hline
FedBN\cite{FedBN} & \begin{tabular}[c]{@{}c@{}}96.6(0.1)\end{tabular} & \begin{tabular}[c]{@{}c@{}}71.0(0.3)\end{tabular} & \textbf{\begin{tabular}[c]{@{}c@{}}97.0(0.3)\end{tabular}} & \begin{tabular}[c]{@{}c@{}}83.2(0.4)\end{tabular} & \begin{tabular}[c]{@{}c@{}}78.3(0.7)\end{tabular} & \begin{tabular}[c]{@{}c@{}}85.2(0.4)\end{tabular} & \begin{tabular}[c]{@{}c@{}}51.2(1.4)\end{tabular} & \begin{tabular}[c]{@{}c@{}}26.8(0.5)\end{tabular} & \begin{tabular}[c]{@{}c@{}}41.5(1.4)\end{tabular} & \begin{tabular}[c]{@{}c@{}}71.3(0.7)\end{tabular} & \begin{tabular}[c]{@{}c@{}}54.8(0.8)\end{tabular} & \begin{tabular}[c]{@{}c@{}}42.1(1.3)\end{tabular} & \begin{tabular}[c]{@{}c@{}}48.0(1.0)\end{tabular} \\ \hline
\textbf{Our RFedDis} & \textbf{\begin{tabular}[c]{@{}c@{}}97.2(0.1)\end{tabular}} & \textbf{\begin{tabular}[c]{@{}c@{}}75.3(0.6)\end{tabular}} & \begin{tabular}[c]{@{}c@{}}96.8(0.2)\end{tabular} & \textbf{\begin{tabular}[c]{@{}c@{}}86.9(0.5)\end{tabular}} & \textbf{\begin{tabular}[c]{@{}c@{}}83.9(1.3)\end{tabular}} & \textbf{\begin{tabular}[c]{@{}c@{}}88.0(0.5)\end{tabular}} & \textbf{\begin{tabular}[c]{@{}c@{}}52.6(1.2)\end{tabular}} & \begin{tabular}[c]{@{}c@{}}27.7(1.1)\end{tabular} & \textbf{\begin{tabular}[c]{@{}c@{}}42.4(0.6)\end{tabular}} & \textbf{\begin{tabular}[c]{@{}c@{}}73.6(0.8)\end{tabular}} & \textbf{\begin{tabular}[c]{@{}c@{}}58.7(1.4)\end{tabular}} & \begin{tabular}[c]{@{}c@{}}43.5(1.2)\end{tabular} & \textbf{\begin{tabular}[c]{@{}c@{}}49.8(1.0)\end{tabular}} \\ 
\toprule[1pt]
\end{tabular}
}
  \label{tab:DigitsAndDomainNet} 
   \vspace{-3pt}
\end{table*}

2) Although our proposed RFedDis does not achieve the highest test accuracy on some clients of Digits and DomainNet, it outperforms SingleSet, which demonstrates that the proposed RFedDis improves the performance of FL. Meanwhile, our proposed RFedDis still achieves better performance than other FL methods on most clients of both datasets, which also demonstrates the robustness of the proposed method. 

3) As shown in Table~\ref{tab:Office-Caltech10} and Table~\ref{tab:DigitsAndDomainNet}, personalized FL methods designed for non-IID label settings, such as FedRep~\cite{FedRep}, FedDC~\cite{FedDC}, and FedDyn~\cite{FedDyn}, etc., perform poorly on non-IID domain feature tasks.
In particular, seen from Table~\ref{tab:Office-Caltech10}, FedProx~\cite{FedProx}, FedRep~\cite{FedRep}, Moon~\cite{Moon}, and FedDyn~\cite{FedDyn} achieve even worse results than SingleSet, due to the fact that these methods try to improve performance at the model level, ignoring exploring non-IID domain features at the feature distribution level, which may lead to introducing domain-specific features in model aggregation to interfere with global generic information learning, resulting in worse performance than SingleSet. In contrast to these personalized FL methods, the performance of our proposed RFedDis is significantly improved, which demonstrates the effectiveness of the core idea of our proposed RFedDis, i.e, introducing disentangled feature learning into the FL to explore non-IID domain features at the feature space level. 

\subsection{Ablation studies}
To further validate the effectiveness and reliability of our proposed RFedDis. We conduct comprehensive ablation studies on the dataset of Office-Caltech10~\cite{OfﬁceCaltech10}. 
Table~\ref{tab:AE} shows the mean (std) of the results of the ablation experiments on Office-Caltech10~\cite{OfﬁceCaltech10}. In this paper, FedBN~\cite{FedBN} without feature disentangling and Uncertainty-aware decision fusion is employed as our backbone. Backbone+$L_{Dis}$ indicates the fusion of decoupled features directly by a simple feature summation operation. As seen in Table~\ref{tab:AE}, Backbone+$L_{Dis}$ achieves better performance than SingleNet due to the benefits of distributed joint training. Moreover, although the introduction of the specific header for feature disentangling increases the network parameters, the direct feature summation leads to the interference of the decoupled domain-specific features with the global generic information, which causes performance degradation compared to the backbone.
Therefore, effective fusion of decoupled features is crucial to improve the performance of FL on non-IID domain features. To this end, we introduce Uncertainty-aware decision fusion to guide the network to dynamically integrate the decoupled features at the evidence level(Backbone+$L_{Dis}$+$L_{Un}$). As shown in Table~\ref{tab:AE}, Backbone+$L_{Dis}$+$L_{Un}$ has significantly improved performance for all clients compared to Backbone and Backbone+$L_{Dis}$, which also demonstrates the effectiveness of introducing Uncertainty-aware fusion for the integration of decoupled features. Finally, compared with Backbone+$L_{Dis}$+$L_{Un}$, the performance of Backbone+$L_{Dis}$+$L_{Un}$+$L_{CE}$, i.e., our proposed RFedDis is further improved significantly, which proves the effectiveness of applying cross-entropy loss({$L_{CE}$}) on the decoupled features of the global and local heads($L_{CE}^L$ +$L_{CE}^G$) to ensure the confidence of the decoupled original features during the iterative fusion optimization. In summary, these ablation experimental results show the effectiveness of the main components in our proposed RFedDis.\parskip=0pt

\begin{table}[!t]
 \captionsetup{font=small}
 \caption{\small{Mean (std) of the results of the ablation experiments on Office-Caltech10.}}
  \vspace{-4pt}
  \centering
\resizebox{0.48\textwidth}{!}{
\begin{tabular}{lll|cccc|c}
\toprule[1pt]
\multicolumn{1}{p{0.5cm}|}{$L_{Dis}$} & \multicolumn{1}{p{0.5cm}|}{$L_{Un}$} & \multicolumn{1}{p{0.5cm}|}{$L_{CE}$} & \multicolumn{1}{c|}{A} & \multicolumn{1}{c|}{C} & \multicolumn{1}{c|}{D} & \multicolumn{1}{c|}{W} & \multicolumn{1}{c}{Avg} \\ \bottomrule[1pt]
\multicolumn{3}{c|}{SingleSet} & \begin{tabular}[c]{@{}c@{}}54.9(1.5)\end{tabular} & \begin{tabular}[c]{@{}c@{}}40.2(1.6)\end{tabular} & \begin{tabular}[c]{@{}c@{}}78.7(1.3)\end{tabular} & \begin{tabular}[c]{@{}c@{}}86.4(2.4)\end{tabular} & \begin{tabular}[c]{@{}c@{}}65.1(1.7)\end{tabular} \\ \hline
 / & / & / & \begin{tabular}[c]{@{}c@{}}63.0(1.6)\end{tabular} & \begin{tabular}[c]{@{}c@{}}45.3(1.5)\end{tabular} & \begin{tabular}[c]{@{}c@{}}83.1(2.5)\end{tabular} & \begin{tabular}[c]{@{}c@{}}90.5(2.3)\end{tabular} & \begin{tabular}[c]{@{}c@{}}70.5(2.0)\end{tabular} \\ \hline
 \checkmark & / & / & \begin{tabular}[c]{@{}c@{}}55.5(2.0)\end{tabular} & \begin{tabular}[c]{@{}c@{}}41.9(1.4)\end{tabular} & \begin{tabular}[c]{@{}c@{}}84.4(2.2)\end{tabular} & \begin{tabular}[c]{@{}c@{}}87.1(3.5)\end{tabular} & \begin{tabular}[c]{@{}c@{}}67.2(2.3)\end{tabular} \\ \hline
 \checkmark & \checkmark & / & \begin{tabular}[c]{@{}c@{}}63.5(1.1)\end{tabular} & \begin{tabular}[c]{@{}c@{}}46.7(1.2)\end{tabular} & \begin{tabular}[c]{@{}c@{}}90.6(2.8)\end{tabular} & \begin{tabular}[c]{@{}c@{}}93.2(1.9)\end{tabular} & \begin{tabular}[c]{@{}c@{}}73.5(1.8)\end{tabular} \\ \hline
 \checkmark & \checkmark & \checkmark & \textbf{\begin{tabular}[c]{@{}c@{}}64.5(1.0)\end{tabular}} & \textbf{\begin{tabular}[c]{@{}c@{}}48.2(0.9)\end{tabular}} & \textbf{\begin{tabular}[c]{@{}c@{}}95.3(3.1)\end{tabular}} & \textbf{\begin{tabular}[c]{@{}c@{}}95.3(2.1)\end{tabular}} & \textbf{\begin{tabular}[c]{@{}c@{}}75.8(1.8)\end{tabular}} \\ \bottomrule[1pt]
\end{tabular}
}
\label{tab:AE}
\end{table}
\begin{table}[!t]
 \vspace{-2pt}
 \captionsetup{font=small}
 \caption{\small{Mean (std) of the test accuracy of updated local epoch \textit{E}.}}
\vspace{-8pt}
  \centering
\resizebox{0.48\textwidth}{!}{
\begin{tabular}{l|cccc|c}
\toprule[1pt]
\multicolumn{1}{l|}{Epoch} & \multicolumn{1}{c|}{A} & \multicolumn{1}{c|}{C} & \multicolumn{1}{c|}{D} & \multicolumn{1}{c|}{W} & Avg \\ \bottomrule[1pt]
1 & \begin{tabular}[c]{@{}c@{}}64.5(1.5)\end{tabular} & \begin{tabular}[c]{@{}c@{}}48.2(0.9)\end{tabular} & \begin{tabular}[c]{@{}c@{}}95.3(3.1)\end{tabular} & \begin{tabular}[c]{@{}c@{}}95.3(2.1)\end{tabular} & \begin{tabular}[c]{@{}c@{}}75.8(1.8)\end{tabular} \\ \hline
4 & \begin{tabular}[c]{@{}c@{}}66.2(1.9)\end{tabular} & \begin{tabular}[c]{@{}c@{}}48.4(2.7)\end{tabular} & \begin{tabular}[c]{@{}c@{}}94.8(1.7)\end{tabular} & \begin{tabular}[c]{@{}c@{}}96.0(2.0)\end{tabular} & \begin{tabular}[c]{@{}c@{}}76.4(1.1)\end{tabular} \\ \hline
16 & \textbf{\begin{tabular}[c]{@{}c@{}}67.4(1.3)\end{tabular}} & \textbf{\begin{tabular}[c]{@{}c@{}}49.3(0.5)\end{tabular}} & \textbf{\begin{tabular}[c]{@{}c@{}}95.8(1.7)\end{tabular}} & \textbf{\begin{tabular}[c]{@{}c@{}}97.2(2.0)\end{tabular}} & \textbf{\begin{tabular}[c]{@{}c@{}}77.0(1.4)\end{tabular}} \\ 
\bottomrule[1pt]
\end{tabular}
}
\label{tab:LocalEpochesTest}
\vspace{-14pt}
\end{table}

In addition, we also conduct experiments to validate the performance of the proposed RFedDis with different locally updated epoch $E\in \left\{1,4,16\right\}$. As shown in Table~\ref{tab:LocalEpochesTest}, the performance of the proposed method is further improved as the number of locally updated epochs increases, and the experimental results further prove the effectiveness of our proposed RFedDis.
\begin{figure*}[!t]
\vspace{9pt}
 \begin{center}
  \includegraphics[width=1\linewidth]{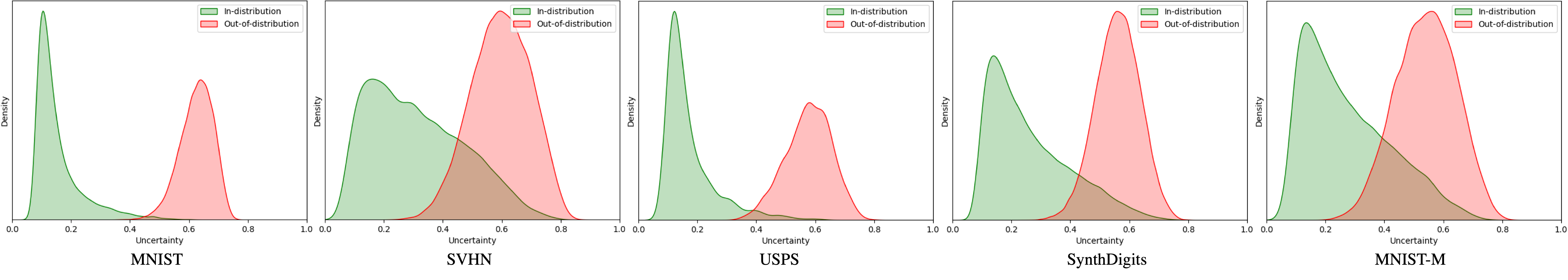}
 \end{center}
 \captionsetup{font=small}
    \vspace{-18pt}
 \caption{\small{Density of uncertainty. Where green represents the density distribution of uncertainty for in-distribution samples, while red is the density distribution of uncertainty for out-of-distribution samples generated by adding Gaussian noise to the original testing data.}}
 \label{DOU}
\end{figure*}

\begin{figure*}[!t]
  \vspace{4pt}
 \begin{center}
  \includegraphics[width=1\linewidth]{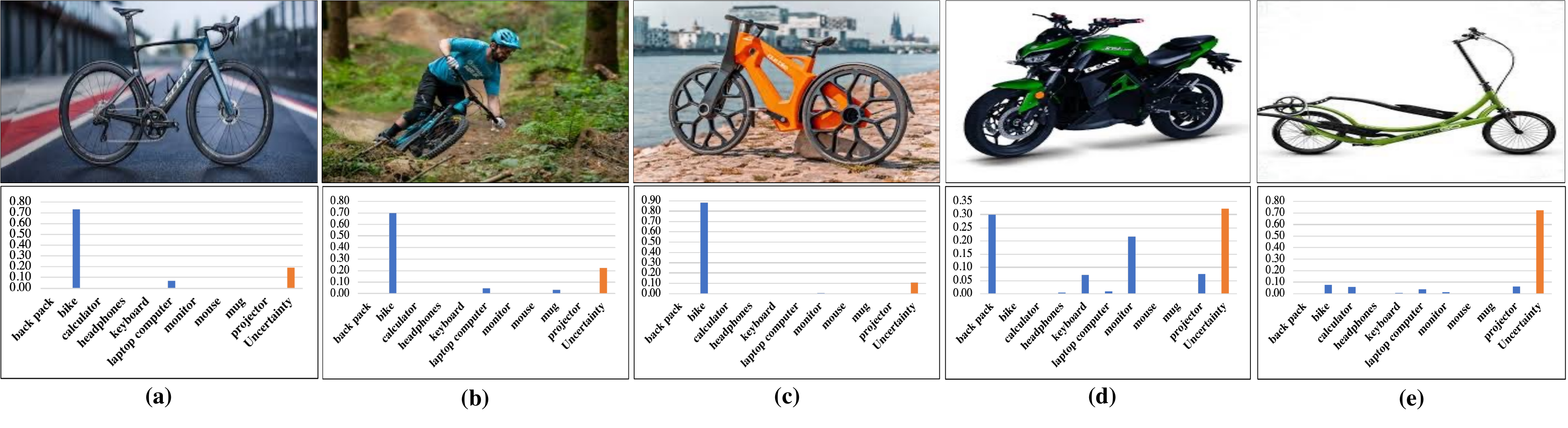}
 \end{center}
    \vspace{-15pt}
 \caption{\small{Reliability analysis. (a-c) Results for high-confidence feature samples with low uncertainty scores. (d-e) Out-of-distribution samples with high uncertainty scores.}}
\vspace{-14pt}
 \label{GoogleImages}
\end{figure*}
\vspace{-10pt}
\subsection{The reliability of our proposed RFedDis} 
To evaluate the reliability of the proposed RFedDis, as shown in Fig.~\ref{DOU}, we visualize the distribution of in-/out- of-distribution samples in terms of uncertainty for different clients on the Digits dataset. Specifically, the original samples are considered as in-distribution data, while the samples generated by adding Gaussian noise ($\sigma =1.5$) to the original input image are used as out-of-distribution data. It can be observed from Fig.~\ref{DOU} that: The in-distribution data with higher accuracy tend to get low uncertainty scores (the green area in Fig.~\ref{DOU}), i.e, the prediction results are reliable. Conversely, out-of-distribution data with lower precision received higher uncertainty scores(the red area in Fig.~\ref{DOU}), i.e, the prediction results may be unreliable. These experimental results show that our proposed RFedDis is a novel reliable FL paradigm as it can provide a reasonable uncertainty evaluation for the prediction result.

Furthermore, we also conduct experiments to further validate the reliability and safety of our proposed RFedDis. As can be seen from Fig.~\ref{GoogleImages}, we downloaded some images through Google that never appeared in the dataset. These images contain samples that are consistent with the training data categories (see Fig.~\ref{GoogleImages} (a-c)), and targets that are not covered in the training dataset labels, i.e., anomalous data (see Fig.~\ref{GoogleImages} (d-e)). The categories of these images were predicted using RFedDis trained on the Office-Caltech dataset. Fig.~\ref{GoogleImages} shows the prediction results with uncertainty evaluation for different samples. As shown in Fig.~\ref{GoogleImages}, the prediction results of the samples with obvious features (see Fig.~\ref{GoogleImages} (a-b)) obtain higher confidence and lower uncertainty scores, which indicates that the prediction results can be reliable. Moreover, although the proposed RFedDis gives incorrect final predictions for the anomalous samples (see Fig.~\ref{GoogleImages} (d-e)), it also gives higher uncertainty scores, indicating that the predictions may not be trustworthy. These experimental results further show that our proposed RFedDis can generate a reliable prediction with an estimated uncertainty, thus potentially avoiding disasters decision from anomalous samples.

\section{Conclusion}
\label{Conclusion}
In this paper, focusing on the non-IID domain feature in the FL task, we propose a novel reliable federated disentangling network (RFedDis). 
The proposed FL paradigm based on feature disentangling enables the network to capture a global generic domain-invariant cross-client representation while enhancing the learning of local domain-specific features. 
Moreover, to effectively fuse the decoupled features, based on the evidential uncertainty, the uncertainty-aware decision fusion is proposed to guide the network to dynamically integrate the decoupled features at the evidence level, while generating an uncertainty score for the final prediction to evaluate its confidence.
Comprehensive experiments on three datasets with non-IID domain feature issues demonstrate that the proposed RFedDis achieves better performance than state-of-the-art FL approaches. 
Meanwhile, unlike previous FL approaches, our RFedDis can generate a reliable prediction with an estimated uncertainty without accuracy loss, which makes the model’s decision more reliable.

\bibliographystyle{IEEEtran}
\bibliography{refs}

\end{document}